\documentclass[10pt,letterpaper]{article}
\usepackage[top=0.85in,left=2.75in,footskip=0.75in]{geometry}

\usepackage{amsmath,amssymb}

\usepackage{changepage}
\usepackage{makecell}

\usepackage[utf8x]{inputenc}

\usepackage{textcomp,marvosym}

\usepackage{cite}

\usepackage{nameref,hyperref}

\usepackage{microtype}
\DisableLigatures[f]{encoding = *, family = * }

\usepackage[table]{xcolor}

\usepackage{array}

\newcolumntype{+}{!{\vrule width 2pt}}

\newlength\savedwidth

\raggedright
\usepackage[aboveskip=1pt,labelfont=bf,labelsep=period,justification=raggedright,singlelinecheck=off]{caption}

\makeatletter
\renewcommand{\@biblabel}[1]{\quad#1.}
\makeatother

\usepackage{lastpage,fancyhdr,graphicx}
\usepackage{epstopdf}
\fancyheadoffset[L]{2.25in}
\fancyfootoffset[L]{2.25in}
\begin{document}
\vspace*{0.2in}

\begin{flushleft}
{\Large
\textbf\newline{Distributed representation of patients and its use for medical risk adjustment} 
}
\newline
\\
Xianlong Zeng\textsuperscript{1},
Soheil Moosavinasab\textsuperscript{2},
En-Ju D Lin\textsuperscript{2},
Razvan Bunescu\textsuperscript{1},
Chang Liu\textsuperscript{1},
Simon Lin\textsuperscript{2},
\\
\bigskip
\textbf{1} School of Electrical Engineering and Computer Science, Ohio University, Athens, OH
\\
\textbf{2} The Research Institute, The Research Institute at Nationwide Children’s Hospital, Columbus, OH
\\
\bigskip

* Corresponding author: liuc@ohio.edu

\end{flushleft}
\section*{Abstract}
Efficient representation of patients is very important in the healthcare domain and can help with many tasks such as medical risk prediction. Many existing methods, such as Diagnostic Cost Groups (DCG), rely on expert knowledge to build patient representation from medical data, which is resource consuming and non-scalable. Unsupervised machine learning algorithms are a good choice for automating the representation learning process. However, there is very little research focusing on patient-level representation learning directly. In this paper, we proposed a novel patient vector learning architecture that learns high quality, fixed-length patient representation from claims data. In addition, our model can learn meaningful medical visit representation and medical code representation at the same time. We conducted several experiments to test the quality of our learned representation, and the empirical results show that our learned patient vectors are superior to vectors learned through other methods. We also used our patient vector on a real-world application, and it outperforms a popular commercial model. Lastly, we provide potential clinical interpretation for using our representation on predictive tasks, as interpretability is vital in the healthcare domain\footnote{The codes for our model is avilable on github: https://github.com/1230pitchanqw/p2v}.

\section*{Introduction}
With the increasing adoption of electronic healthcare records (EHR), more healthcare and medical data are available digitally. The large amount of health data offers opportunities to apply machine learning methods in many predictive healthcare tasks. Some algorithms, such as logistic regression, typically prefer the input feature vectors to be small and efficient. One basic way to represent the medical coding data is through a bag-of-words (BOW) like approach: each medical code can be represented as a one-hot vector and each patient can be represented by aggregating the one-hot medical code vectors. However, representing the patients in such a high-dimensional and sparse vector will not only lose the temporal and co-occurrent information between medical visits and within medical codes, but also make it difficult for the machine learning algorithms to make a stable prediction without overfitting for different medical outcomes.
\\
Medical concepts, especially medical codes, are not independent from each other. For example, \textit{Type 2 diabetes mellitus} (ICD-9 250.00) is obviously more related to \textit{Type 1 diabetes mellitus} (ICD-9 250.01) than \textit{Asthma, unspecified} (ICD-9 439.00). Representing a medical code in a one-hot vector will not capture such relationships. To address this representation issue, medical experts developed many models that can group related medical codes for different purposes. For example, Clinical Classifications Software (CCS) is a tool developed for grouping related diagnosis and procedure codes into a manageable number (around 600) of clinically meaningful categories~\cite{bib1,bib2,bib3}. Diagnostic Cost Groups (DCG) group the diagnosis codes from medical risk perspectives, and it is able to calculate patients’ expected future medical costs based on diagnoses, age and gender~\cite{bib4,bib5,bib6,bib7,bib8}. DCG has been implemented by commercial vendors like Truven Health Analytics (Ann Arbor, Michigan). Although models based on domain knowledge provide us with a good way to capture the relationships between medical concepts, they are very resource consuming and labor-intensive to develop, maintain and update.
\\
To overcome this limitation, many researchers utilized machine learning methods to learn efficient representations of medical concepts without relying on domain knowledge. One possible way is through supervised learning, and the representation will be learned as a “side effect.” Choi et al.~\cite{bib9} built a recurrent neural network (RNN) model to predict patients’ future diagnoses. Baytas and colleagues~\cite{bib10} built a time-aware RNN model to handle time irregularities in medical visit sequences. Choi et al.~\cite{bib11} used an interpretable attention based multi-layer perceptron (MLP) model to predict different medical outcomes. Ma et al.~\cite{bib12} developed a bidirectional attention RNN model to predict future medical codes.  All these models are supervised, and efficient representations of patients are learned automatically to achieve a good predictive outcome. However, the quality of the representation is not the major focus for supervised learning tasks, and the representation learning process might be biased for certain kinds of predictive tasks.
\\
Another common way to learn high quality representation is unsupervised learning. Choi et al.~\cite{bib13} and Choi et al.~\cite{bib14} developed models based on skip-gram to learn medical codes representation. Miotto et al.~\cite{bib15} and Baytas et al.~\cite{bib10} learned patient representation via auto-encoder (AE) based models. Skip-gram~\cite{bib18} based models are able to capture the co-occurrence information within medical visits. However, the existing skip-gram based models are only able to learn representation for medical codes and medical visits. In order to obtain patient representation, one needs to aggregate the code/visit vectors by averaging or adding, which will lose the temporal order of information in medical visits. RNN-AE based models~\cite{bib10} are able to capture the temporal relationship of the medical visits. However, RNN based models are notoriously hard to train when the length of medical sequences is long. In addition, results are not interpretable.
\\
To address the aforementioned representation learning challenges in healthcare, we proposed Patient Vector (PV), an unsupervised framework that learns continuous distributed vector representation for patients. Our model is inspired by the recent work in learning paragraph vectors from English documents~\cite{bib16} and learning medical visit vectors from EHR~\cite{bib13}. Our main contributions are listed below:
\\
\begin{enumerate}
  \item We proposed a novel architecture to learn patient representation that can be used for many predictive tasks including medical cost prediction. Moreover, our model is able to learn a meaningful representation for medical visits and medical codes.
  \item We demonstrated our learned patient representation improves predictive model performance in various healthcare tasks compared to other baselines. Additionally, our model outperforms another popular commercial model in practical use.
  \item We showed the potential interpretability of our learned representation by conducting several case studies and analyzing the clinical meaning of our representation.
\end{enumerate}

\section*{Materials and methods}
\subsection*{Dataset}

Our dataset contains medical claims between 2014 to 2016 for over 300,000 members  from Partners for Kids (PFK), one of the largest Accountable Care Organizations (ACO’s) for low-income children in central and southeastern Ohio. In accordance with the Common Rule (45 CFR 46.102[f]) and the policies of Nationwide Children’s Institutional Review Board, this study used a limited data set and was not considered human subjects research and thus not subject to institutional review board approval. 
\\
We extracted members’ medical information from their medical claims including: 1. Visit level information, where each visit contains the claims within the same service date, and it includes medical codes (diagnosis, procedure, medication) and utilization (cost, place of service, category of the visit). 2. Individual level information (age, sex, annual cost). Note that the "cost" we refer here and the rest of the paper denote the actual amount that is paid to the provider. 
\\
We removed members without continuous eligibility across 2014 to 2016 and members who have less than two medical visits during 2014 to 2015. This reduced the size of our cohort from over 300,000 to 107,060 members. We hereby refer to this group of members as patients. We acknowledge this eligibility requirement creates a biased sample as it removes those members that might be experiencing more instability (inferred by their inconsistent eligibility within these 3 years). However, without continuous eligibility, we cannot ascertain if the annual medical cost is accurate. For the purpose of evaluating the models, we have chosen to focus on this particular sample.
\\
The detailed statistical information for our cohort is presented in Table~\ref{table1}. We trained our model on 2014-2015 data and evaluated the trained embedding on 2016 data. The details of the evaluation metric will be provided in the following sections.
\begin{table}[!ht]
\centering
\caption{
{\bf  Statistics information of our dataset.}}
\begin{center}
\begin{tabular}{ |c|c| } 
 \hline
\# of members &107,060 \\ 
\# of visits &1,547,471 \\
Avg. \# of visits per member &14.5 \\ 
Avg. \# of codes per member & 37.7 \\
\hline
\# of unique medical codes & 13,620\\
\# of unique diagnosis codes & 8,164\\ 
\# of unique medication codes &339 \\ 
\# of unique procedure codes & 5,117\\
\hline
Max \# of codes per member & 347\\
(95\%, 99\%) \# of codes per member  &(81,118)\\
(95\%, 99\%) \# of codes per visit & (10,15) \\
\hline
\end{tabular}
\end{center}
\label{table1}
\end{table}

\subsection*{Data preprocessing}
Due to the ICD-9 to ICD-10 conversion in Oct, 2015, there were inconsistent diagnosis codes in our dataset. To address the mapping inconsistency, we converted the ICD-10 codes to ICD-9 codes using the General Equivalence Mappings (GEMS) publicly available mapping table~\cite{bib17}. We ignored the codes that could not be mapped in GEMS. 
\\
Our dataset contains medical visits (claims for medical procedures) and retail pharmacy claims. In order to train all medical codes (diagnosis, procedure, medication) under the same latent space, we needed to combine the pharmacy claims and medical visits. To do so, we used the service date to track member's medical visits. Since the service dates for pharmacy claims could be a few days after the corresponding medical visit, we combined the pharmacy claim and medical visit if the pharmacy claim occurred within two weeks after a medical visit. We ignored the pharmacy claims if there was no medical visit before 2 weeks.
\\
Lastly, as some members have more than one insurance, their medical claim might have been paid by another insurer, resulting in a non-positive paid amount in some instances. This occurred in approximately 4\% of such medical claims in our dataset, and we converted the paid amount in such claims to zero.

\subsection*{Basic Notation}
We denote all the unique medical codes from the dataset as $C =\{c_1, c_2, ..., c_{|C|}\}$, where $C$ is the vocabulary of all medical codes. Claims data for each patient contains $T$ medical visits $V_1, V_2, ..., V_T$, where $V_t$ contains a subset of medical codes and ordered by time-stamp $t$. Initially, we represent each patient by a count vector $y_i \in \{0,1,2,...,k\}^{|C|}$, where the $j^{th}$ index is $k$ if $p^i$ contains code $c_j$ $k$ times across all medical visits, visit $V_t$ is represented by a binary vector $x_t \in \{0,1\}^{|C|}$, where $j^{th}$ column is 1 if $V_t$ contains code $c_j$. Our dataset contains three types of medical codes: diagnosis, encoded by International Classification of Diseases (ICD); procedure, encoded by Current Procedural Terminology (CPT); and medication, encoded by drug classes.
\\
\subsection*{Learning from code-level co-occurrence information} 
Medical codes within each medical visit contain co-occurrence information: related medical codes are likely to share similar set of codes as context~\cite{bib13,bib14}. One common way to capture such relationship is via Skip-gram, a model proposed by Miklov et al.~\cite{bib18} that is able to capture the syntactic and semantic relationships of English words. The main idea of Skip-gram is to use a word to predict its neighboring words in a sentence. By doing so, words sharing similar context will have a similar representation in the latent space, we can also use the Skip-gram model to capture the syntactic and semantic relationship of medical codes, as done by Choi et al.~\cite{bib14} and Choi et al.~\cite{bib13}. More formally, given a sequence of medical codes $c_1, c_2, ..., c_{i}$ within a given medical visit, our objective function is to maximize the average log likelihood:
$$ \frac{1}{T}\sum_{t=1}^{T} \sum_{-w<j<w,j\ne0} \log P(c_{t+j} | c_t)$$
, where $w$ is the context window size and the probability is computed using softmax function:

$$ P(c_o | c_i) = \frac{exp(v_{c_o}^T v_{c_i})}{\sum_{j=1, |c|} exp(v_j^Tv_{c_i})}$$
, where $v_c$ is the representation for code $c$. Note that we do not distinguish between the “input” and “output” medical code as suggested by Choi et al.~\cite{bib13} considering the unordered structure of medical codes within the medical visit.
\\
\subsection*{Learning from visit-level sequential information}
In Med2vec~\cite{bib13}, the model proposed by Choi et al., visit vectors are formed by summing the code vectors and used to predict the codes in neighbouring visits in order to capture the sequential information between medical visits. Similar to Med2vec, we created the same medical visits prediction task but with a small modification. Inspired by the idea from Paragraph-vector architecture~\cite{bib16}, we created an additional patient vector $P^{(i)}$ for each patient and asked it to contribute to predicting the medical codes within the neighboring medical visits. The patient vector can be thought of as the generalization of patient’s health condition. It acts as a memory that remembers the patient’s overall medical history, and contributes to the prediction of surrounding visits. The patient vector helped the current visit to remember what else it needs when predicting the surrounding visits. The patient vector is shared across all medical visit prediction tasks for the same patient.
\\
More formally, as shown in Fig~\ref{fig1}, given a patient $p^i$ and his/her medical visits $V_1, V_2, ..., V_T$, the count patient vector $y_i$ is converted into an intermediate patient representation $k_i$ and the binary visit vector $x^i_t$ is converted into an intermediate visit representation $u^i_t$ via the following equations:
$$u_t^i = W_c x_t^i + b_c$$
$$ k^i = W_c y^i + b_c $$
, where $W_c$, $b_c$ are the code weight matrix and bias. Then we concatenate the patient-level demographic information $d^i$ and visit-level demographic information $d_t^i$ separately to create the final patient representation $p^i$ and visit representation $v_t^i$ as follows:
$$v_t^i = ReLU(W_v[u_t^i, d_t^i] + b_v) $$
$$p^i = ReLU(W_p[k^i, d^i]+ b_p) $$
, where $W_v$, $b_v$ are the visit weight matrix and bias, $W_p$, $b_p$ are the patient weight matrix and bias. We use ReLU as the activation function. Lastly, we concatenate the patient representation and visit representation to predict the medical codes of the visits within a fixed context window via a softmax model:
$$ \hat x_t = softmax(W_o[p^i, v_t^i] + b_o ) $$
, where $W_v$, $b_v$ are the output weight matrix and bias.
Our objective is to minimize the following cross entropy function:
$$ \frac{1}{T}\sum_{t=1}^{T} \sum_{-w<j<w,j\ne0} -x_{t+j}^T log \hat{x}_{t+j} -(1-x_{t+j})^T log(1- \hat{x}_{t+j})$$

\begin{figure}[!h]
\caption{{\bf Patient Vector model architecture.}
The training objectives are to: 1) learn medical code representation that is good at predicting neighboring codes within the same visit. 2) learn medical visit representation and patient representation that are good at predicting nearby visits.}
\centering
\label{fig1}
\end{figure}

\subsection*{Unified training}
To capture both code-level co-occurrence information as well as visit-level sequential information, we combined the aforementioned two objective functions together using a hyper-parameter $\lambda$ as follows:
$$ \frac{1}{T}\sum_{t=1}^{T} \sum_{-w<j<w,j\ne0} -x_{t+j}^T log \hat{x}_{t+j} -(1-x_{t+j})^T log(1- \hat{x}_{t+j})$$
\\
$$+ \lambda \frac{1}{T}\sum_{t=1}^{T} \sum_{-w<j<w,j\ne0} \log P(c_{t+j} | c_t)$$

\subsection*{From classification to ranking}
One drawback of our Patient Vector model is the big softmax matrix $W_o$ at the output layer. To alleviate the curse of dimensionality when training the softmax classifier, Med2vec~\cite{bib13} utilized the hierarchical structure of medical codes and trained the classifier to predict the grouped medical codes instead of the exact medical codes. By doing so, the final output space can be significantly reduced from around 20,000 to 2,000 dimensions. However, the output dimension is still very high and the curse of dimensionality problem remains unsolved.
\\
To better alleviate this issue, we proposed a different objective function denoted as score ranking function. We denote the Patient Vector model with score ranking objective function as Patient Vector+, as shown in Fig~\ref{fig2}.

\begin{figure}[!h]
\caption{{\bf Patient Vector+ model architecture.}
The training objective is: 1) learn medical code representation that are good at predicting codes within the same visit. 2) learn medical visit representation and patient representation that will give high score for codes within nearby visits.}
\centering
\label{fig2}
\end{figure}

In the Patient Vector+ framework, instead of calculating the probability of all possible medical codes in the neighbor visits, we randomly select $k$ negative codes (i.e., codes that are not in the neighbor visits) and train the model to assign low scores for these negative codes, while for positive codes (codes that are in the neighbor visits), the model should assign high scores.
\\
More formally, we will use the new objective function as below:
$$\frac{1}{T} \sum_{t=1}^{T} \sum_{-w<j<w,j\ne0}\sum_{c_m \in x_{t+j}^i} \sum_{c_{n} \notin x_{t+j}^i} max(0, \gamma-score(x_t^i,y^i, c_m)+score(x_t^i,y^i, c_{n}))$$\\
$$ + \lambda \frac{1}{T} \sum_{t=1}^{T}\sum_{-w<j<w,j\ne0}logP(c_{t+j}|c_t)$$
\\
, where
$$ score(x_t^i,y^i, c) =W_p[p_i, V_t^i, W_c[:,c]]+ b_p $$

\subsection*{Implementation detail}
All the approaches were implemented using Python and Tensorflow with the dataset randomly divided into training, validation and testing set in 7:1:2 ratio by patient. For the training process, we set minibatch size = 100, visit window size $w=1, \lambda=\{0.5,1\}, k=\{10,20\}$, patient embedding size=\{100,200\}, visit embedding size=\{100,200\} and code embedding size=\{100, 200\}.  We performed 40 training epochs over the training dataset, and select the hyperparameters based on the model performance on validation dataset. We used early stopping method on validation dataset to prevent overfitting, and report the model performance on testing dataset.

\section*{Results and discussion}
\subsection*{Patient representation evaluation}
To evaluate the quality of our patient representation, we first compared our model with other baselines for two different predictive tasks: current year cost prediction and next year cost prediction. Then, we applied our model to a real-world application: high medical risk patient selection, and compared our model with DCG, a popular commercial model.

\begin{itemize}
    \item {\bf Task 1: Current year cost prediction}. The medical cost of current year is based on patients’ medical utilization information including medical procedures, medical utilization and medication usage. One way to evaluate the learned patient representation is to use it for predicting the current year’s medical cost. In our current cost prediction task, we used the learned patient vector as the input feature and current medical cost (from 2014 to 2015) as the output label to train a regression model.
    \item {\bf Task 2: Next year cost prediction}. The next year medical cost of patients is very important for hospitals and ACO’s. With more accurate prediction of future medical cost, hospitals and ACO’s can do better financial forecast and financial risk management. In our future cost prediction task, we used learned patient vector and previous medical cost (from 2014 to 2015) as input features, and the future medical cost (in 2016) as the output label to train a regression model.
    \item {\bf Task 3: High medical risk patient selection}. With the shift to managed healthcare for patients, ACO’s and providers are responsible for managing the overall health of a patient. Many hospitals and ACO’s have care coordination programs aimed to improve health outcomes and reduce medical costs for patients with high medical risk. The DCG model is a population-based classification and risk adjustment methodology that is widely used by organizations and hospitals to evaluate a patient’s future medical risk.  The DCG model assigns a score for each patient with a higher score indicating higher predicted medical costs in the next year. In the high medical risk patient selection task, we compared patients with the highest predicted medical cost by DCG and our model. 
\end{itemize}
To compare our model with other baselines, we used the Coefficient of determination ($R^2$) and Root Mean Squared Error (RMSE) evaluation metrics for task-1 and task-2. Note that we calculated the below two values on the log-scaled medical cost in order to normalize the highly skewed distribution of medical costs, as suggest by Diehr et al~\cite{bib19}. Five different random seed were randomly seleted and used to split the dataset. We reported the average results for each model, and denote standard deviations within the parentheses.

\begin{itemize}
    \item {\bf Coefficient of determination ($R^2$)} is the standard approach for the evaluation of regression models. It represents the proportion of the variance in the output value that is predictable from the input variables. The best possible value for $R^2$ is 1, when the output value can be predicted 100\% from the input variables. 
    \item  {\bf Root Mean Squared Error (RMSE)} is also a commonly used measurement for regression models. It measure the differences between predicted values and the true values. 
\end{itemize}
The following baseline methods are used in comparison.
\begin{itemize}
    \item {\bf Count vector model}: Patients are represented by a count vector, where element at position $i$ represents the number of occurrences for the $i^{th}$ code in vocabulary.
    \item {\bf Stacked denoising auto-encoder (SDA)}: A three layer SDA model to learn the patient representation as described by Miotto et al.~\cite{bib15}.
    \item {\bf Sum skip-gram vectors (Skip-gram)}: Patients are represented by aggregating their medical code vectors, which are learned by the skip-gram model as described in previous studies~\cite{bib13,bib14}
    \item {\bf Sum Med2vec visit vectors (Med2vec)} Patients are represented by aggregating their visit vectors, which are learned by the Med2vec~\cite{bib13} model.
    \item {\bf Concatenation of visit and code vectors (Med2vec+)}: To enhance the previous two baselines, we concatenated the visit and code embeddings to form patient embedding.
\end{itemize}
 Table~\ref{table2} shows the evaluation results for different patient embedding learning approaches. Our patient vector models generally outperforms the other models. In particular, the Patient Vector+ model consistently performed the best out of all the models evaluated. Raw count vector model performed worst for cost prediction tasks as the input feature space is too high for a regression model to make a meaningful prediction.
\begin{table}[!ht]
\centering
\caption{
{\bf  Evaluation of patient embedding on two medical risk prediction tasks.} Model with higher $R^2$ value and lower RMSE value indicate the model is better fitting the task.}
\begin{center}
\begin{tabular}{ |c|c|c| } 
 \hline
Models  & \makecell{Task-1 Result \\ ($R^2$,RMSE)} &\makecell{Task-2 Result \\ ($R^2$,RMSE)}\\ 
\hline
Raw count vector & 15.66\%(0.0857), 1.086(0.0711) & 15.03\%(0.0231), 1.765(0.0605) \\
\hline
SDA & 28.47\%(0.0022), 1.071(0.0066) & 26.09\%(0.0037), 1.650(0.0134) \\
\hline
Skip-gram & 42.36\%(0.0052), 0.900(0.0054) & 26.44\%(0.0034), 1.645(0.0122)\\ 
\hline
Med2vec (13) & 58.22\%(0.0056), 0.766(0.0064) &  28.39\%(0.0045), 1.623(0.0119) \\
\hline
Med2vec+  & 61.11\%(0.0090), 0.739(0.0061) &  28.66\%(0.0041), 1.620(0.013) \\
\hline
Patient Vector & 57.34\%(0.0048), 0.780(0.0045) &28.55\% (0.0049), 1.622 (0.0134) \\
\hline
{\bf Patient Vector+} & {\bf 66.75\%(0.0048), 0.684(0.0075)} &{\bf  29.88\%(0.0035), 1.611(0.0122)}\\ 
\hline
\end{tabular}
\end{center}
\label{table2}
\end{table}

Fig~\ref{fig3} shows the empirical results for the high medical risk patient selection task. We selected patients with the top 0.5\%, 1\%, 5\%, 10\% and 50\% highest predicted medical cost by DCGs and our model. Selected patients with higher future medical cost (in 2016) indicate a better fitting model. For the top 0.5\% predicted high risk patient group, patients selected by the Patient Vector models had  much higher medical costs in the predicted year (in 2016), indicating our model was able to more accurately identify the subset of the most expensive patients. These high cost patients are likely to spend lots of medical resources in the future. Prospective identification of these patients will allow hospitals and ACO’s to provide interventions such as care coordination.

\begin{figure}[!h]
\caption{{\bf High cost patient selection by three different models.} Model with higher value on y-axis indicate the model is better at selecting future high cost patients.}
\centering
\label{fig3}
\end{figure}

\subsection*{Visit representation evaluation}
To evaluate the quality of our visit representation, we used our learned visit representation for similar tasks as patient representation evaluation tasks: current visit cost prediction and next visit cost prediction.
\\
Table~\ref{table3}  shows the evaluation results for visit embedding. Med2vec~\cite{bib13} model is doing better than our methods in current visit cost prediction task and worse than our model in next visit cost prediction task. From these two tasks, we know that our methods are able to learn efficient representation for medical visits compared to the previous state-of-art medical visit representation learning method.
\begin{table}[!ht]
\centering
\caption{
{\bf  Evaluation of visit embedding.} Model with higher $R^2$ value and lower RMSE value indicate the model is better fitting the task}
\begin{center}
\begin{tabular}{ |c|c|c| } 
 \hline
Models  & \makecell{Current visit cost prediction
 \\ ($R^2$,RMSE)} &\makecell{Next visit cost prediction \\ ($R^2$,RMSE)}\\ 
\hline
Med2vec (13) &  {\bf41.38\%, 0.930} & 27.19\%, 1.046 \\
\hline
Patient Vector  & 38.85\%,0.949 &  27.29\%, 1.044\\
\hline
Patient Vector+ & 38.81\%,0.955 & {\bf 28.91\%, 1.033} \\
\hline
\end{tabular}
\end{center}
\label{table3}
\end{table}

\subsection*{Code representation evaluation}
{\bf Evaluation by existing medical grouper.} We evaluated the quality of our learned diagnosis code vectors via an existing medical grouper: Clinical Classifications Software (CCS). CCS is a medical grouper that can group diagnosis codes into around 300 different categories based on expert opinion: codes within the same categories are believed to have certain relationship with each other. We plot all the diagnosis codes in Fig~\ref{fig4} using t-SNE projection, and highlighted the diagnosis codes that belong to the same  CCS category in Diamond with the same color, for a subset of the CCS categories. Diagnosis codes from the remaining categories are shown in dim round circle.

\begin{figure}[!h]
\caption{{\bf Diagnosis in t-SNE projected vector space.} Diagnosis code belong to the same CCS category are closed to each other, indicating the grouping the codes similar to the medical experts.}
\centering
\label{fig4}
\end{figure}

As shown in Fig~\ref{fig4}, most of the diagnosis codes that belong to the same CCS category are likely to be grouped in the same area, such as Asthma (CCS 128) and Ear disorders (CCS 94). However, not all diagnosis codes that belong to the same CCS category are well grouped. For example, we can see some superficial injury (CCS 239) related codes that are located far away from the majority codes. This, according to our observation, is because of two reasons: 1) Artifact caused by dimensionality reduction algorithm. For some separated codes, such as some of the superficial injury related codes we mentioned above, they are actually very close to each other in the original embedding space if we use cosine similarity to measure the distance. It is the t-SNE algorithm that separate them away from the majority. 2) There are some other hidden relationship that is captured. One major idea of our embedding learning algorithm is to capture the co-occurrence information within medical visit: codes with similar neighbors are more close to each other. Hence the learned code representation does not necessarily follow the CCS categorization. For example, Essential hypertension (CCS-98) and Diabetes mellitus (CCS 49)  belong to different CCS categories, but since they often appear together within the same medical visit along with similar diabete/hypertension related procedures and medications, their learned embedding vectors will be close to each other according to our algorithm, which also make sense from the clinical perspective.
\\
Medical groupers such as CCS can be used to evaluate the performance of the learned medical vectors as it can give us a good sense of the overall quality of the embeddings. Based on the Fig~\ref{fig4}, we believed our learned code representation is in high quality.

\subsection*{Interpretability Analysis}
{\bf Interpretation.} Interpretability is very important in healthcare domain. As used in many medical representation learning related papers, we interpret the clinical meaning of each dimension of the learned medical code vectors by selecting the top eight codes that have the largest values for each dimension~\cite{bib13}\cite{bib12}. More formally, to evaluate the clinical meaning of the $i^{th}$ column of the code vector, we selected medical codes via the following equation:
$$ argsort (W_c[:,i][1:8]) $$
, where $W_c[:,i]$  represents the $i^{th}$ column of the code embedding matrix.
\\
Together with analysing the coordinate of code embedding, we are also interested in finding the most influential factors used by the predictive model to make predictions. We analyze the regression model and the clinical meaning of embeddings to find out which code coordinate plays an important role in prediction of current annual cost. We used the analytical method proposed by Choi et al.~\cite{bib19}. The idea is to find out the code coordinates that can maximizes the output activation via the following equation:
$$ k^i = argmax(ReLU(W_pk^i + b_c)W_{LR} *max(W_c+b_c))$$
, where $W_LR$  is the weights of the regression model and we are using broadcasting addition for $W_c+b_c$.
\\
Using the above strategy, we selected the top two coordinates that have the strongest influence and their corresponding eight codes for the current cost prediction, as shown in Table~\ref{table4}. For coordinate 128, it groups medical codes that are related to emergency services and accidents that cause wounds. Coordinate 134 is related to acute diseases such as vomiting and fever, radiology examination and emergency visits. The medical codes associated with the two coordinates are obviously very expensive for children and hence it make sense for our regressor to assign more weights to these two coordinates. This computational result also confirms several currently used strategy to control the cost of pediatric ACO’s: 1) to proactively reduce the usage of emergency rooms, 2) to prevent home accidents, and 3) to control operating room cost.

\begin{table}[!ht]
\caption{
{\bf Medical codes with the maximum value in each coordinate.}}
\begin{center}
\begin{tabular}{ |p{6cm}|p{6cm}|} 
 \hline
Coordinate 128  & Coordinate 134\\
\hline

\tiny 1.EMERGENCY DEPARTMENT VISIT FOR THE EVALUATION A MANAGEMENT OF A (PROC) & \tiny1. FEVER, UNSPECIFIED (78060)
\\
\tiny2.EMERGENCY ROOM - GENERAL CLASSIFICATION (PROC) &\tiny 2.RADIOLOGIC EXAMINATION, CHEST, 2 VIEWS
\\
\tiny3.SERVICE(S) PROVIDED BETWEEN 10:00 PM AND 8:00 AM AT 24-HOUR FACILITY, (PROC) &\tiny3.FRONTAL AND LATERAL; (PROC)
\\
\tiny4.OPERATING ROOM SERVICES-GENERAL CLASSIFICATION (PROC)
 &\tiny4.SERVICE(S) PROVIDED BETWEEN 10:00 PM AND 8:00 AM AT 24-HOUR FACILITY, (PROC)
\\
\tiny5.EMERGENCY DEPARTMENT VISIT FOR THE EVALUATION AND MANAGEMENT OF A (PROC)
 &\tiny5.EMERGENCY ROOM - GENERAL CLASSIFICATION (PROC)
\\
\tiny6.EMERGENCY DEPARTMENT VISIT FOR THE EVALUATION AND MANAGEMENT OF A (PROC)
 &\tiny6.EMERGENCY DEPARTMENT VISIT FOR THE EVALUATION AND MANAGEMENT OF A (PROC)
\\
\tiny7.SIMPLE REPAIR OF SUPERFICIAL WOUNDS OF FACE, EARS, EYELIDS, NOSE, (PROC) &\tiny7.EMERGENCY DEPARTMENT VISIT FOR THE EVALUATION AND MANAGEMENT OF A (PROC)
\\
\tiny8.HOME ACCIDENTS (E8490)
 &\tiny8.VOMITING ALONE (78703)
 \\
\hline
\end{tabular}
\end{center}
\label{table4}
\end{table}

{\bf Patient Similarity.} Once we have learned the patient representation, we can calculate the distance between patients. To analyze whether it makes sense to evaluate the patient similarity via patient embedding, we performed two clustering tasks. Firstly we selected 500 patients with three different diseases (asthma, depression, seizures). Since the three diseases have distinct etiology, patients with these diseases are likely to be separated in their clinical representation. Secondly, we selected 1,500 patients that have either the highest medical cost or the lowest medical cost in the subsequent year. We hypothesize that patients with high future medical cost would have a more complex medical condition than patients with low future medical cost, and the difference could be captured by patient embedding.
\\
As shown in Fig~\ref{fig5}, there are obviously three clusters for patients with different diseases. The clusters are not perfectly separated, this is likely because patient usually have more than one diseases, which will complicated their health condition and affect the embedding vectors. 

\begin{figure}[!h]
\caption{\bf Patient representation plot for patients with three different diseases.}
\centering
\label{fig5}
\end{figure}

We can obviously observe two clusters for patients with different future medical cost from Fig~\ref{fig6}. Most of the high future cost patients are clustered tightly with each other. In contrast, patients with low future cost are loosely spread on the plot with some mixing with the high future cost patients. We postulate that this is due to the heterogeneous nature of acute conditions frequently encountered in children. On the other hand, patients with the highest future cost are more likely to have chronic, severe diseases, that require long-term extensive medical care.

\begin{figure}[!h]
\caption{\bf Patient representation plot for patients with different future medical cost.}
\centering
\label{fig6}
\end{figure}

\section*{Limitation and future studies}
The model we proposed in this study is data-driven and excluded expert knowledge, this is a limitation under the current scope as taking into account domain knowledge is likely to improve the predictive power of healthcare models~\cite{bib20, bib21, bib22}.  Another limitation is we didn't consider the heterogeneity (i.e. different subgroups might have different relationship between variables, like diagnosis, and medical cost). "Average effects" might lead to an inaccurate prediction~\cite{bib23, bib24}.

Although the only predictive output of our model is the paid amount to the healthcare provider, we do understand that it is not the only valuable outcome. Our future work will focus not only on the medical cost, but also the quality and necessity of the medical care.

\section*{Conclusion}
In this paper, we propose a novel architecture to address the challenges of modeling patient representation from medical data. By utilizing a multilayer neural network, we built an unsupervised learning algorithm that learns patient representations that can capture both medical code co-occurrence information and medical visit sequential information. Our learned patient embeddings show superior predictive power when compared with existing patient representation methods and also a commercially used model developed by medical experts. In addition, we also demonstrated the clinical interpretability by applying patient embedding to a real world predictive task.

\section*{Acknowledgments}
We would like to thank Jennifer Klima and Brad Stamm from Partners For Kids (PFK) for providing the medical claims data for this study and for valuable discussions. We also thank Stephen Cardamone and Deena Chisolm for providing useful feedback to the manuscript.

%
%
%

\end{document}